\documentclass{article} 
\usepackage{nips15submit_e,times}
\usepackage{hyperref}
\usepackage{url}

\usepackage{amsmath}
\usepackage{dsfont, bbm}
\usepackage{graphicx}
\usepackage{amsfonts}
\usepackage{amsthm, mathabx}
\usepackage[shortlabels]{enumitem}

\newcommand\numberthis{\addtocounter{equation}{1}\tag{\theequation}}

\newtheorem{thm}{Theorem}
\newtheorem{lem}{Lemma}
\newcounter{factnum}
\newcounter{claimnum}
 \newcounter{defnum}
\usepackage{caption}
\newenvironment{changemargin}[2]{%
  \begin{list}{}{%
    \setlength{\topsep}{0pt}%
    \setlength{\leftmargin}{#1}%
    \setlength{\rightmargin}{#2}%
    \setlength{\listparindent}{\parindent}%
    \setlength{\itemindent}{\parindent}%
    \setlength{\parsep}{\parskip}%
  }%
  \item[]}{\end{list}}
  
\usepackage{multibib} 
\newcites{si}{Additional References for the Supplementary Information}

\newcommand{\beginsupplement}{ 
        \setcounter{section}{0}
        \renewcommand{\thesection}{S\arabic{section}} %
         \renewcommand{\thesubsection}{\thesection.\arabic{subsection}}
        \setcounter{table}{0}
        \renewcommand{\thetable}{S\arabic{table}} %
        \setcounter{figure}{0}
        \renewcommand{\thefigure}{S\arabic{figure}} %
     }

\newcommand{\fullmethod}{principal differences analysis}
\newcommand{\capsmethod}{Principal Differences Analysis}
\newcommand{\methodabbrev}{PDA}
\newcommand{\sparsemethod}{SPARDA}
\newcommand{\relaxed}{RELAX} 

\title{\capsmethod : Interpretable Characterization of Differences between Distributions}

\author{
Jonas Mueller \\
CSAIL, MIT\\
\texttt{jonasmueller@csail.mit.edu} \\
\And
Tommi Jaakkola \\
CSAIL, MIT\\
\texttt{tommi@csail.mit.edu}
}

\nipsfinalcopy 

\begin{document}
\maketitle
\begin{abstract}
We introduce principal differences analysis (\methodabbrev{}) for analyzing differences between high-dimensional distributions. The method operates by finding the projection that maximizes the Wasserstein divergence between the resulting univariate populations. Relying on the Cramer-Wold device, it requires no assumptions about the form of the underlying distributions, nor the nature of their inter-class differences. A sparse variant of the method is introduced to identify features responsible for the differences. We provide algorithms for both the original minimax formulation as well as its semidefinite relaxation.  In addition to deriving some convergence results, we illustrate how the approach may be applied to identify differences between cell populations in the somatosensory cortex and hippocampus as manifested by single cell RNA-seq. Our broader framework extends beyond the specific choice of Wasserstein divergence.
\end{abstract}

\section{Introduction}  
Understanding differences between populations is a common task across disciplines, from biomedical data analysis to demographic or textual analysis. For example, in biomedical analysis, a set of variables (features) such as genes may be profiled under different conditions (e.g.\ cell types, disease variants), resulting in two or more populations to compare. The hope of this analysis is to answer whether or not the populations differ and, if so, which variables or relationships contribute most to this difference. In many cases of interest, the comparison may be challenging primarily for three reasons: 1) the number of variables profiled may be large, 2) populations are represented by finite, unpaired, high-dimensional sets of samples, and 3) information may be lacking about the nature of possible differences (exploratory analysis).  

We will focus on the comparison of two high dimensional populations. Therefore, given two unpaired i.i.d.\ sets of samples $\mathbf{X}^{(n)}= x^{(1)}, \dots, x^{(n)} \sim \mathbb{P}_X$ and $\mathbf{Y}^{(m)} = y^{(1)}, \dots, y^{(m)} \sim \mathbb{P}_Y$, the goal is to answer the following two questions about the underlying multivariate random variables $X, Y \in \mathbb{R}^d$: (Q1) Is $\mathbb{P}_X = \mathbb{P}_Y$? \ (Q2) If not, what is the minimal subset of features $S \subseteq \{1,\dots, d \}$ such that the marginal distributions differ $\mathbb{P}_{X_{S}} \neq \mathbb{P}_{Y_{S}}$ while $\mathbb{P}_{X_{S^C}} \approx \mathbb{P}_{Y_{S^C}}$ for the complement? A finer version of (Q2) may additionally be posed which asks how much each feature contributes to the overall difference between the two probability distributions (with respect to the given scale on which the variables are measured).

Many two-sample analyses have focused on characterizing limited differences such as mean shifts \cite{Lopes2011,Clemmensen2011}. More general differences beyond the mean of each feature remain of interest, however, including variance/covariance of demographic statistics such as income. It is also undesirable to restrict the analysis to specific parametric differences, especially in exploratory analysis where the nature of the underlying distributions may be unknown. In the univariate case, a number of nonparametric tests of equality of distributions are available with accompanying concentration results \cite{Vaart1996}. Popular examples of such \emph{divergences} (also referred to as probability metrics) include: $f$-divergences (Kullback-Leibler, Hellinger, total-variation, etc.), the Kolmogorov distance, or the Wasserstein metric \cite{Gibbs2002}.  Unfortunately, this simplicity vanishes as the dimensionality $d$ grows, and complex test-statistics have been designed to address some of the difficulties that appear in high-dimensional settings \cite{Wei2015, Rosenbaum2005, Szekely2004, Gretton2012}.  

In this work, we propose the \emph{\fullmethod{}} (\methodabbrev{}) framework which circumvents the curse of dimensionality through explicit reduction back to the univariate case. Given a pre-specified statistical divergence $D$ which measures the difference between univariate probability distributions, \methodabbrev{}  seeks to find a projection $\beta$ which maximizes $D(\beta^T X, \beta^T Y)$ subject to the constraints $|| \beta ||_2 \le 1, \beta_1 \ge 0$ (to avoid underspecification).  This reduction is justified by the Cramer-Wold device, which ensures that $\mathbb{P}_X \neq \mathbb{P}_Y$ \emph{if and only if} there exists a direction along which the univariate linearly projected distributions differ \cite{Cramer1936,Cuesta2007,Jirak2011}.  Assuming $D$ is a \emph{positive definite} divergence (meaning it is nonzero between any two distinct univariate distributions), the projection vector produced by \methodabbrev{} can thus capture arbitrary types of differences between high-dimensional $\mathbb{P}_X$ and $\mathbb{P}_Y$.  Furthermore, the approach can be straightforwardly modified to address (Q2) by introducing a sparsity penalty on $\beta$ and examining the features with nonzero weight in the resulting optimal projection. The resulting comparison pertains to marginal distributions up to the sparsity level. We refer to this approach as \emph{sparse differences analysis} or \sparsemethod{}.  

\section{Related Work}
The problem of characterizing differences between populations, including feature selection, has received a great deal of study  \cite{Clemmensen2011, tibshirani1996regression, Bradley1998, Wei2015, Lopes2011}. We limit our discussion to projection-based methods which, as a family of methods, are closest to our approach. For multivariate two-class data, the most widely adopted methods include (sparse) linear discriminant analysis (LDA) \cite{Clemmensen2011} and the logistic lasso \cite{tibshirani1996regression}. While interpretable, these methods seek specific differences (e.g., covariance-rescaled average differences) or 
operate under stringent assumptions (e.g., log-linear model). In contrast, \sparsemethod{} (with a positive-definite divergence) aims to find features that characterize a priori unspecified differences between general multivariate distributions. 


Perhaps most similar to our general approach is Direction-Projection-Permutation (DiProPerm) procedure of Wei et al.\ \cite{Wei2015}, in which the data is first projected along the normal to the separating hyperplane (found using  linear SVM, distance weighted discrimination, or the centroid method) followed by a univariate two-sample test on the projected data. The projections could also be chosen at random \cite{Lopes2011}. In contrast to our approach, the choice of the projection in such methods is not optimized for the test statistics. We note that by restricting the divergence measure in our technique, methods such as the (sparse) linear support vector machine \cite{Bradley1998} could be viewed as special cases. The divergence in this case would measure the margin between projected univariate distributions. While suitable for finding well-separated projected populations, it may fail to uncover more general differences between possibly multi-modal projected populations.

\section{General Framework for \capsmethod}

For a given divergence measure $D$ between two univariate random variables, we find the projection $\widehat{\beta}$ that solves 
\begin{equation}
\max_{\beta \in \mathcal{B}, ||\beta||_0 \le k}\big\{ \ D(\beta^T \widehat{X}^{(n)}, \beta^T \widehat{Y}^{(m)} ) \big\}
\label{generaldpp} 
\end{equation} 
where $\mathcal{B} := \{ \beta \in \mathbb{R}^d : ||\beta||_2 \le 1, \beta_1 \ge 0 \}$ is the feasible set, $||\beta||_0 \le k$ is the sparsity constraint, and $\beta^T \widehat{X}^{(n)}$ denotes the observed random variable that follows the empirical distribution of $n$ samples of $\beta^T X$. Instead of imposing a hard cardinality constraint $||\beta||_0 \le k$, we may instead penalize by adding a penalty term\footnote{In practice, shrinkage parameter $\lambda$ (or explicit cardinality constraint $k$) may be chosen via cross-validation by maximizing the divergence between held-out samples.}  $-\lambda ||\beta||_0$ or its natural relaxation, the $\ell_1$ shrinkage used in Lasso \cite{tibshirani1996regression}, sparse LDA \cite{Clemmensen2011}, and sparse PCA  \cite{DAspremont2007, Amini2009}.  Sparsity in our setting explicitly restricts the comparison to the marginal distributions over features with non-zero coefficients. We can evaluate the null hypothesis $\mathbb{P}_X = \mathbb{P}_Y$ (or its sparse variant over marginals) using permutation testing (cf. \cite{Wei2015, Good1994}) with statistic $D(\widehat{\beta}^T \widehat{X}^{(n)}, \widehat{\beta}^T \widehat{Y}^{(m)} )$. 

The divergence $D$ plays a key role in our analysis. If $D$ is defined in terms of density functions as in $f$-divergence, one can use univariate kernel density estimation to approximate projected pdfs with additional tuning of the bandwidth hyperparameter. For a suitably chosen kernel (e.g.\ Gaussian), the unregularized \methodabbrev{} objective (without shrinkage) is a smooth function of $\beta$, and thus amenable to the projected gradient method (or its accelerated variants \cite{Duchi2011,Wright2010}). In contrast, when $D$ is defined over the cdfs along the projected direction -- e.g.\ the Kolmogorov or Wasserstein distance that we focus on in this paper -- the objective is nondifferentiable due to the discrete jumps in the empirical cdf. We specifically address the combinatorial problem implied by the Wasserstein distance. Moreover, since the divergence assesses general differences between distributions, Equation (\ref{generaldpp}) is typically a non-concave optimization. To this end, we develop  a semi-definite relaxation for use with the Wasserstein distance. 

\section{\methodabbrev{} using the Wasserstein Distance}

In the remainder of the paper, we focus on the squared $L_2$ Wasserstein distance (a.k.a. Kantorovich, Mallows, Dudley, or earth-mover distance), defined as
\begin{equation}
D(X, Y) = \min_{\mathbb{P}_{XY}} \mathbb{E}_{\mathbb{P}_{XY}} || X - Y ||^2 \ \ \text{ s.t. } \ (X,Y) \sim \mathbb{P}_{XY}, \ X \sim \mathbb{P}_X, \ Y \sim \mathbb{P}_Y
\label{wassdef}
\end{equation}
where the minimization is over all joint distributions over $(X,Y)$ with given marginals $\mathbb{P}_X$ and $\mathbb{P}_Y$.  Intuitively interpreted as the amount of \emph{work} required to transform one distribution into the other, 
$D$ provides a natural dissimilarity measure between populations that integrates both the fraction of individuals which are different and the magnitude of these differences.  
While component analysis based on the Wasserstein distance has been limited to \cite{Sandler2011}, this divergence has been successfully used in many other applications \cite{Levina2001}. In the univariate case, (\ref{wassdef}) may be analytically expressed  as the $L_2$ distance between quantile functions.  We can thus efficiently compute empirical projected Wasserstein distances by sorting $X$ and $Y$ samples along the projection direction to obtain quantile estimates. 

Using the Wasserstein distance, the empirical objective in Equation (\ref{generaldpp}) between unpaired sampled populations $\{x^{(1)},\ldots,x^{(n)}\}$ and $\{y^{(1)},\ldots,y^{(m)}\}$ can be shown to be 
\begin{equation}
\max_{\substack{\beta \in \mathcal{B} \\ ||\beta||_0 \le k}} \bigg\{ \min_{M \in \mathcal{M}, } \ \sum_{i=1}^n \sum_{j = 1}^m (\beta^T x^{(i)} -  \beta^T y^{(j)})^2 M_{ij} \bigg\}
= \max_{\substack{\beta \in \mathcal{B} \\ ||\beta||_0 \le k}} \left\{ \min_{M \in \mathcal{M}} \  \beta^T W_M \beta \right\}
\label{wassdpp} 
\end{equation}
where $\mathcal{M}$ is the set of all $n \times m$ nonnegative \emph{matching} matrices with fixed row sums $=1/n$ and column sums $=1/m$ (see \cite{Levina2001} for details), $W_M := \sum_{i,j} [Z_{ij} \otimes Z_{ij}] M_{ij}$, and $Z_{ij} := x^{(i)} - y^{(j)}$. If we omitted (fixed) the inner minimization over the matching matrices and set $\lambda=0$, the solution of (\ref{wassdpp}) would be simply the largest eigenvector of $W_M$. Similarly, for the sparse variant without minizing over $M$, the problem would be solvable as sparse PCA \cite{DAspremont2007, Amini2009, Wang2014}.  The actual max-min problem in (\ref{wassdpp}) is more complex and non-concave with respect to $\beta$. We propose a two-step procedure similar to ``tighten after relax'' framework used to attain minimax-optimal rates in sparse PCA \cite{Wang2014}. First, we first solve a convex relaxation of the problem and subsequently run a steepest ascent method (initialized at the global optimum of the relaxation) to greedily improve the current solution with respect to the original nonconvex problem whenever the relaxation is not tight. 

Finally, we emphasize that \methodabbrev{} (and \sparsemethod{}) not only computationally resembles (sparse) PCA, but the latter is actually a special case of the former in the Gaussian, paired-sample-differences setting.  This connection is made explicit by considering the two-class problem with \emph{paired} samples $(x^{(i)}, y^{(i)})$ where $X, Y$ follow two multivariate Gaussian distributions.  Here, the largest principal component of the (uncentered) differences $x^{(i)} - y^{(i)}$ is in fact equivalent to the direction which maximizes the projected Wasserstein difference between the distribution of $X - Y$ and a delta distribution at 0.

\subsection{Semidefinite Relaxation}
The \sparsemethod{} problem may be expressed in terms of $d \times d$ symmetric matrices $B$ as  
\begin{align*}
& \max_B\ \min_{M \in \mathcal{M}} \ \text{tr}\left(  W_M B \right) \\
& \text{subject to} \;\; \text{tr}(B) = 1,\;
B \succeq 0, \; ||B||_0 \le k^2,  \; \text{rank}(B) = 1
\numberthis \label{rewrittendpp}
\end{align*}
where the correspondence between (\ref{wassdpp}) and (\ref{rewrittendpp}) comes from writing $B =  \beta \otimes \beta$ (note that any solution of (\ref{wassdpp}) will have unit norm).  When $k = d$, i.e., we impose no sparsity constraint as in \methodabbrev{}, we can relax by simply dropping the rank-constraint. The objective is then a supremum of linear functions of $B$ and the resulting semidefinite problem is concave over a convex set and may be written as:
\begin{equation}
\max_{B \in \mathcal{B}_r} \ \  \min_{M \in \mathcal{M}} \ \  \text{tr}\left(  W_M B \right) 
  \label{relaxation}
\end{equation}
where $\mathcal{B}_r$ is the convex set of positive semidefinite $d \times d$ matrices with trace = 1. If $B^* \in \mathbb{R}^{d\times d}$ denotes the global optimum of this relaxation and rank$(B^*) = 1$, then the best projection for \methodabbrev{} is simply the dominant eigenvector of $B^*$ and the relaxation is tight. Otherwise, we can truncate $B^*$ as in \cite{DAspremont2007}, treating the dominant eigenvector as an approximate solution to the original problem (\ref{wassdpp}).  

To obtain a relaxation for the sparse version where $k < d$ (\sparsemethod{}), we follow \cite{DAspremont2007} closely.   Because $B = \beta \otimes \beta$ implies $||B||_0 \le k^2$, we can thus get an equivalent cardinality constrained problem by incorporating this nonconvex constraint into (\ref{rewrittendpp}).  Since \text{tr}$(B) = 1$ and $||B||_F = ||\beta||_2^2=1$,  a convex relaxation of the squared $\ell_0$ constraint is given by $||B||_1 \le k$.  By selecting $\lambda$ as the optimal Lagrange multiplier for this $\ell_1$ constraint, we can obtain an equivalent penalized reformulation parameterized by $\lambda$ rather than $k$ \cite{DAspremont2007}.  The sparse semidefinite relaxation is thus the following concave problem
\begin{equation}
\max_{B \in \mathcal{B}_r} \big\{\ \min_{M \in \mathcal{M}} \ \text{tr}\left(  W_M B \right)  - \lambda || B ||_{1} \ \big\}
  \label{sparserelaxation}
\end{equation}
While the relaxation bears strong resemblance to DSPCA relaxation for sparse PCA, the inner maximization over matchings prevents direct application of general semidefinite programming solvers. Let $M(B)$ denote the matching that minimizes  $\text{tr}\left(  W_M B \right)$ for a given $B$. Standard projected subgradient ascent could be applied to solve (\ref{sparserelaxation}), where at the $t^{\text{th}}$ iterate the (matrix-valued) subgradient would be given by $W_{M(B^{(t)})}$. However, this approach requires us to maintain large $n\times m$ matrices. Instead, we resort to the dual (cf. \cite{Bertsekas1998, Bertsekas1988})  
\[
\min_{M \in \mathcal{M}} \ \text{tr}\left(  W_M B \right)
= \frac{1}{m}\max_{u,v} \sum_{i=1}^n\sum_{j=1}^m \min\{0, \ \text{tr}([Z_{ij} \otimes Z_{ij}]\,B)-u_i-v_j\} + \frac{1}{n}\sum_{i=1}^n u_i + \frac{1}{m}\sum_{j=1}^m v_j
\]
that enables us to alternatingly maximize (\ref{sparserelaxation}) over $B\in \mathcal{B}_r$,  $u\in \mathbb{R}^n$, and $v\in \mathbb{R}^m$  
without requiring matching matrices nor their cumbersome row/column constraints.  While $u$ and $v$ can be solved in closed form for each fixed $B$ (via sorting), we describe a simple sub-gradient approach that works better in practice.

\noindent\rule[0.5ex]{\linewidth}{1pt}
\textbf{RELAX Algorithm:} Solves the dualized semidefinite relaxation of \sparsemethod{} (\ref{sparserelaxation}).  Returns the largest eigenvector of the solution to (\ref{sparserelaxation})  as the desired projection direction for \sparsemethod{}.  \\
\noindent\rule[0.5ex]{\linewidth}{1pt}
\textbf{Input:} $d$-dimensional data $x^{(1)}, \dots, x^{(n)}$ and $y^{(1)}, \dots, y^{(m)}$ (with $n \ge m$) \\
\textbf{Parameters:} $\lambda \ge 0$ controls the amount of regularization, $\gamma > 0$ is the step-size used for $B$ updates, $\eta > 0$ is the step-size used for updates of dual variables $u$ and $v$, $T$ is the maximum number of iterations without improvement in cost after which algorithm terminates.
\begin{enumerate}[1:, topsep=-1.5ex,leftmargin=6mm]
\item Initialize $\beta^{(0)} \leftarrow \left[\frac{\sqrt{d}}{d},\dots, \frac{\sqrt{d}}{d} \right]$, $B^{(0)} \leftarrow \beta^{(0)}  \otimes \beta^{(0)} \in \mathcal{B}_r$, $u^{(0)} \leftarrow \mathbf{0}_{n \times 1}$, $v^{(0)}\leftarrow \mathbf{0}_{m \times 1}$
\item \textbf{While} the number of iterations since last improvement in objective function is less than $T$:
\item  \hspace*{4ex} $\partial u \leftarrow \left[1/n , \dots, 1/n  \right] \in \mathbb{R}^n$, \ $\partial v \leftarrow \left[1/m , \dots, 1/m  \right] \in \mathbb{R}^m$, \ $\partial B \leftarrow \mathbf{0}_{d \times d}$
\item  \hspace*{4ex} \textbf{For}  $i,j \in \{1,\dots,n\} \times \{1,\dots,m\}$: \ \ \ 
\item  \hspace*{8ex} $Z_{ij} \leftarrow x^{(i)} - y^{(j)}$
\item  \hspace*{8ex} \textbf{If} \ $\text{tr}([Z_{ij} \otimes Z_{ij}] \hspace*{0.2mm} B^{(t)})-u^{(t)}_i-v^{(t)}_j < 0$  :
\item \hspace*{12ex} $\partial u_i \leftarrow \partial u_i - 1/m$\hspace*{0.8mm}, \ $\partial v_j \leftarrow \partial v_j - 1/m$\hspace*{0.8mm}, \ $ \partial B \leftarrow \partial B + Z_{ij} \otimes Z_{ij} \hspace*{0.3mm} / m$
\item   \hspace*{4ex}  \textbf{End For}
\item  \hspace*{4ex} $u^{(t+1)} \leftarrow u^{(t)} + \eta \cdot  \partial u$ \ and \ $v^{(t+1)} \leftarrow v^{(t)}  + \eta \cdot \partial v$
\item \hspace*{4ex} $B^{(t+1)} \leftarrow$ \textbf{Projection}$\left(B^{(t)} + \frac{\gamma }{ ||\partial B||_F}   \cdot \partial B \ ; \ \lambda \hspace*{0.2mm}, \ \gamma / ||\partial B||_F \right)$
\end{enumerate}
\textbf{Output:} $\widehat{\beta}_{\text{relax}} \in \mathbb{R}^d$ defined as the largest eigenvector (based on corresponding eigenvalue's magnitude) of the matrix $B^{(t^*)}$ which attained the best objective value over all iterations. \\
\noindent\rule[0.5ex]{\linewidth}{1pt}

\clearpage
\noindent\rule[0.5ex]{\linewidth}{1pt}
\textbf{Projection Algorithm:} Projects matrix onto positive semidefinite cone of unit-trace matrices $\mathcal{B}_r$ (the feasible set in  our relaxation).   Step 4 applies soft-thresholding proximal operator for sparsity. \\  
\noindent\rule[0.5ex]{\linewidth}{1pt}
\textbf{Input:} $B \in \mathbb{R}^{d \times d}$ \\
\textbf{Parameters:} $\lambda \ge 0$ controls the amount of regularization, $\delta = \gamma / ||\partial B||_F  \ge 0$ is the actual step-size used in the $B$-update.  
\begin{enumerate}[1:, topsep=-1.5ex,leftmargin=6mm]
\item $Q \Lambda Q^T \leftarrow$ eigendecomposition of  $B$
\item $\displaystyle w^* \leftarrow \arg\min \left\{ ||w - \text{diag}(\Lambda)||_2^2 : w \in [0,1]^d , ||w||_1 = 1 \right\}$ \hspace*{18mm} (Quadratic program)
\item $\widetilde{B} \leftarrow Q \cdot  \text{diag}\{w^*_1,\dots,w^*_d \} \cdot Q^T$
\item \textbf{If} $\lambda > 0$:  \ \textbf{For} $r, s \in \{1,\dots, d\}^2:    \hspace*{5mm}  \widetilde{B}_{r,s} \leftarrow \text{sign}(\widetilde{B}_{r,s}) \cdot \max\{ 0, | \widetilde{B}_{r,s}  | -  \delta \lambda  \}$
\end{enumerate}
\textbf{Output:} $\widetilde{B} \in \mathcal{B}_r $ \\
\noindent\rule[0.5ex]{\linewidth}{1pt}

The \relaxed{} algorithm (boxed) is a projected subgradient method with supergradients computed in Steps 3 - 8. For scaling to large samples, one may alternatively employ \emph{incremental} supergradient directions \cite{Bertsekas2011} where Step 4 would be replaced by drawing random $(i,j)$ pairs.  After each subgradient step, projection back into the feasible set $\mathcal{B}_r$ is done via a quadratic program involving the current solution's eigenvalues. In \sparsemethod{}, sparsity is encouraged via the soft-thresholding proximal map corresponding to the $\ell_1$ penalty. The overall form of our iterations matches  subgradient-proximal updates (4.14)-(4.15) in \cite{Bertsekas2011}.  By the convergence analysis in  \S4.2 of \cite{Bertsekas2011}, the RELAX algorithm (as well as its incremental variant) is guaranteed to approach the optimal solution of the dual which also solves (\ref{sparserelaxation}), provided we employ sufficiently large $T$ and small step-sizes.  In practice, fast and accurate convergence is attained by: (a) renormalizing the $B$-subgradient (Step 10) to ensure balanced updates of the unit-norm constrained $B$, (b) using diminishing learning rates which are initially set larger for the unconstrained dual variables (or even taking multiple subgradient steps in the dual variables per each update of $B$).

\subsection{Tightening after relaxation}

It is unreasonable to expect that our semidefinite relaxation is always tight. Therefore, we can sometimes further refine the projection $\widehat{\beta}_{\text{relax}}$ obtained by the RELAX algorithm by using it as a starting point in the original non-convex optimization. We introduce a sparsity constrained \emph{tightening} procedure for applying projected gradient ascent for the original nonconvex objective $J(\beta) = \min_{M \in \mathcal{M}} \beta^T W_M \beta$ where $\beta$ is now forced to lie in $\mathcal{B} \cap \mathcal{S}_k$ and $\mathcal{S}_k := \{ \beta \in \mathbb{R}^d : ||\beta||_0 \le k \}$. The sparsity level $k$ is fixed based on the relaxed solution ($k = ||\widehat{\beta}_{\text{relax}}||_0$). After initializing $\beta^{(0)} =  \widehat{\beta}_{\text{relax}} \in \mathbb{R}^d$, the tightening procedure iterates steps in the gradient direction of $J$ followed by straightforward projections into the unit half-ball $\mathcal{B}$ and the set $\mathcal{S}_k$ (accomplished by greedily truncating all entries of $\beta$ to zero besides the largest $k$ in magnitude).

Let $M(\beta)$ again denote the matching matrix chosen in response to $\beta$. $J$ fails to be differentiable at the $\widetilde{\beta}$ where $M(\widetilde{\beta})$ is not unique. This occurs, e.g., if two samples have identical projections under $\widetilde{\beta}$.  While this situation becomes increasingly likely as $n, m \rightarrow \infty$, $J$ interestingly becomes smoother overall (assuming the   distributions admit density functions). For all other $\beta$: $M(\beta') = M(\beta)$ where $\beta'$ lies in a small neighborhood around $\beta$ and $J$ admits a well-defined gradient $2W_{M(\beta)} \beta$. In practice, we find that the tightening always approaches a local optimum of $J$ with a diminishing step-size. We note that, for a given projection, we can efficiently calculate gradients without recourse to matrices $M(\beta)$ or $W_{M(\beta)}$ by sorting ${\beta^{(t)}}^T x^{(1)}, \dots, {\beta^{(t)}}^T x^{(n)}$ and ${\beta^{(t)}}^T y^{(1)}, \dots, {\beta^{(t)}}^T y^{(m)}$. The gradient is directly derivable from expression (\ref{wassdpp}) where the nonzero $M_{ij}$ are determined by appropriately matching empirical  quantiles (represented by sorted indices) since the univariate Wasserstein distance is simply the $L_2$ distance between quantile functions \cite{Levina2001}.  Additional computation can be saved by employing insertion sort which runs in nearly linear time for almost sorted points (in iteration $t-1$, the points have been sorted along  the $\beta^{(t-1)}$ direction and their sorting in direction $\beta^{(t)}$ is likely similar under small step-size).  Thus the tightening procedure is much more efficient than the RELAX algorithm (respective runtimes are $O(d \hspace*{0.3mm} n\log n)$ vs.\ $O(d^3 n^2)$ per iteration). 

We require the combined steps for good performance.  The projection found by the tightening algorithm heavily depends on the starting point $\beta^{(0)}$, finding only the closest local optimum (as in \ Figure \ref{spardaresults}a).  It is thus important that $\beta^{(0)}$ is already a good solution, as can be produced by our RELAX algorithm.  Additionally, we note that as first-order methods, both the RELAX and tightening algorithms are amendable to a number of (sub)gradient-acceleration schemes (e.g.\ momentum techniques,  adaptive learning rates, or FISTA and other variants of Nesterov's method \cite{Wright2010, Duchi2011, Beck2009}).  

\subsection{Properties  of semidefinite relaxation} \label{relaxation-property}
We conclude the algorithmic discussion by highlighting basic conditions under which our \methodabbrev{} relaxation is tight.  Assuming $n, m \rightarrow \infty$, each of (i)-(iii) implies that the $B^*$ which maximizes (\ref{relaxation}) is nearly rank one, or equivalently $B^* \approx \widetilde{\beta} \otimes  \widetilde{\beta}$ (see Supplementary Information \S\ref{relaxderivations} for intuition).   Thus, the tightening procedure initialized at $\widetilde{\beta}$ will produce a global maximum of the \methodabbrev{} objective.
\vspace*{-6mm}
\begin{enumerate}[(i)]
\item There exists direction in which the \emph{projected} Wasserstein distance between $X$ and $Y$ is nearly as large as the overall Wasserstein distance in $\mathbb{R}^d$.  This occurs for example if $|| \mathbb{E}[X] - \mathbb{E}[Y] ||_2$ is large while both $||\text{Cov}(X)||_F$ and $||\text{Cov}(Y)||_F$ are small (the distributions need not  be Gaussian). 
\item  $X \sim N(\mu_X, \Sigma_X)$ and $Y  \sim N(\mu_Y, \Sigma_Y)$ with $\mu_X \neq \mu_Y$ and $\Sigma_X \approx \Sigma_Y$.
\item  $X \sim N(\mu_X, \Sigma_X)$ and $Y  \sim N(\mu_Y, \Sigma_Y)$ with $\mu_X = \mu_Y$ where the underlying covariance structure is such that $\arg\max_{B \in \mathcal{B}_r} || (B^{1/2}  \Sigma_X B^{1/2} )^{1/2} - (B^{1/2} \Sigma_Y B^{1/2})^{1/2}  ||_F^2$ is nearly rank 1.  For example, if the primary difference between covariances is a shift in the marginal variance of some features, i.e.\ $\Sigma_Y \approx V \cdot  \Sigma_X$ where $V$ is a diagonal matrix.  
\end{enumerate}

\section{Theoretical Results}

In this section, we characterize statistical properties of an empirical divergence-maximizing projection $\displaystyle \widehat{\beta} := \arg \max_{\beta \in \mathcal{B}} \ D(\beta^T \widehat{X}^{(n)}, \beta^T \widehat{Y}^{(n)} )$, although we note that the algorithms may not succeed in finding such a global maximum for severely nonconvex problems.  Throughout, $D$ denotes the squared $L_2$ Wasserstein distance between univariate distributions, $C$ represents universal constants that change from line to line.  All proofs are relegated to the Supplementary Information \S\ref{sec:proofs}.  We make the following simplifying assumptions: (A1) $n=m$  \ (A2) $X, Y$ admit continuous density functions \ (A3) $X,Y$ are compactly supported with nonzero density in the Euclidean ball of radius $R$.  Our theory can be generalized beyond (A1)-(A3) to obtain similar (but  complex) statements through careful treatment of the distributions'  tails and zero-density regions where cdfs are flat.
\begin{thm} Suppose there exists direction $\beta^* \in \mathcal{B}$ such that $D({\beta^*}^T X ,  {\beta^*}^T Y ) \ge \Delta$.  Then:
\begin{equation*}
 D(\widehat{\beta}^T \widehat{X}^{(n)}, \widehat{\beta}^T \widehat{Y}^{(n)} ) > \Delta - \epsilon  \ \ \ \ \text{ with probability greater than } 1 - 4 \exp \left( -\frac{n \epsilon^2}{16 R^4} \right)
\end{equation*}
\label{directionthm}
\end{thm}
Theorem  \ref{directionthm} gives basic concentration results for the projections used in empirical applications our method.  To relate distributional differences between $X, Y$ in the ambient $d$-dimensional space with their estimated divergence along the univariate linear representation chosen by \methodabbrev{}, we turn to Theorems \ref{equalitythm} and \ref{multivariatethm}.  Finally, Theorem \ref{sparsistency} provides sparsistency guarantees for \sparsemethod{}  in the case where $X, Y$ exhibit large differences over a certain feature subset (of known cardinality).
\begin{thm}
If $X$ and $Y$ are identically distributed in $\mathbb{R}^d$, then: \ $D(\widehat{\beta}^T \widehat{X}^{(n)}, \widehat{\beta}^T \widehat{Y}^{(n)} ) < \epsilon$ \\
with probability greater than 
\[   1 - C_1  \left(1 +  \frac{R^2}{\epsilon}\right)^d \exp\left(-  \frac{C_2}{R^4} n \epsilon^2 \right) 
\]
\label{equalitythm}
\end{thm}
\vspace*{-2mm}
To measure the difference between the untransformed random variables $X, Y$ $\in \mathbb{R}^d$, we define the following metric between distributions on $\mathbb{R}^d$ which is parameterized by $a \ge 0$ (cf.\ \cite{Jirak2011}): 
\begin{align}
&  T_a(X, Y) := {\large|} \Pr (| X_1 | \le a, \dots, |X_d| \le a) -   \Pr (| Y_1 | \le a, \dots, |Y_d| \le a){\large|} \label{ambientmetric}
\end{align}

In addition to (A1)-(A3), we assume the following for the next two theorems:  (A4) $Y$ has sub-Gaussian tails, meaning cdf $F_Y$ satisfies: $1-F_Y(y) \le \frac{C}{y} \exp(-y^2 / 2)$, \ (A5) $\mathbb{E}[X] = \mathbb{E}[Y] = 0$ (note that mean differences can trivially be captured by linear projections, so these are not the differences of interest in the following theorems), (A6) Var($X_\ell$) = 1 for $\ell = 1,\dots,d$
\begin{thm} Suppose $\exists \ a \ge 0$ s.t.\ $T_a( X ,  Y ) > h\left(g(\Delta) \right)$ where $ \ h\left( g(\Delta)\right) :=  \min \{\Delta_1 , \Delta_2 \}$ with 
\begin{align}
& \Delta_1 :=  (a + d)^d (g(\Delta) + d) + \exp(-a^2/2) + \psi \exp\left(-1/(\sqrt{2}\psi) \right) \\
& \Delta_2 := \left(g(\Delta)  + \exp (-a^2/ 2) \right) \cdot d
\end{align}
 $\psi := || \text{Cov}(X) ||_1$, \ $g(\Delta) :=  \Delta^4 \cdot \left( 1 +\Phi  \right)^{-4} $, \ and $\Phi := \sup_{\alpha \in \mathcal{B}} \ \big\{ \sup_y |f_{\alpha^T Y}(y) | \big\}$ \\
with $f_{\alpha^T Y}(y)$ defined as the density of the projection of $Y$ in the $\alpha$ direction. \\
Then:
\begin{equation} D(\widehat{\beta}^T \widehat{X}^{(n)}, \widehat{\beta}^T \widehat{Y}^{(n)} ) > C \Delta - \epsilon
\label{quantcw}
\end{equation}
with probability greater than $1 - C_1 \exp \left( -\frac{C_2}{R^4} n  \epsilon^2 \right)$
\label{multivariatethm}
\end{thm}

\begin{thm} Define $C$ as in (\ref{quantcw}).  Suppose there exists feature subset $S \subset \{1,\dots, d\}$ s.t.\ $|S| = k$,  $T( X_S ,  Y_S ) \ge  h\left( g \left( \epsilon (d + 1)/ C \right) \right)$, and remaining marginal distributions $X_{S^C}$ ,  $Y_{S^C}$ are identical.   Then:
$$ \widehat{\beta}^{(k)} := \arg \max_{\beta \in \mathcal{B}} \ \{ D(\beta^T \widehat{X}^{(n)}, \beta^T \widehat{Y}^{(n)} )  : || \beta ||_0 \le k \}  
$$
satisfies $\widehat{\beta}^{(k)}_i \neq 0$ and $\widehat{\beta}^{(k)}_j = 0 \  \ \forall \ i \in S, j \in S^C$ with probability greater than 
$$ 1 - C_1  \left(1 +  \frac{R^2}{\epsilon}\right)^{d-k} \exp\left(-  \frac{C_2}{R^4} n \epsilon^2 \right) $$
\label{sparsistency}
\end{thm} 
\vspace*{-9mm}

\section{Experiments}
\begin{figure}[b!] \centering 
 \includegraphics[width=0.3\textwidth]{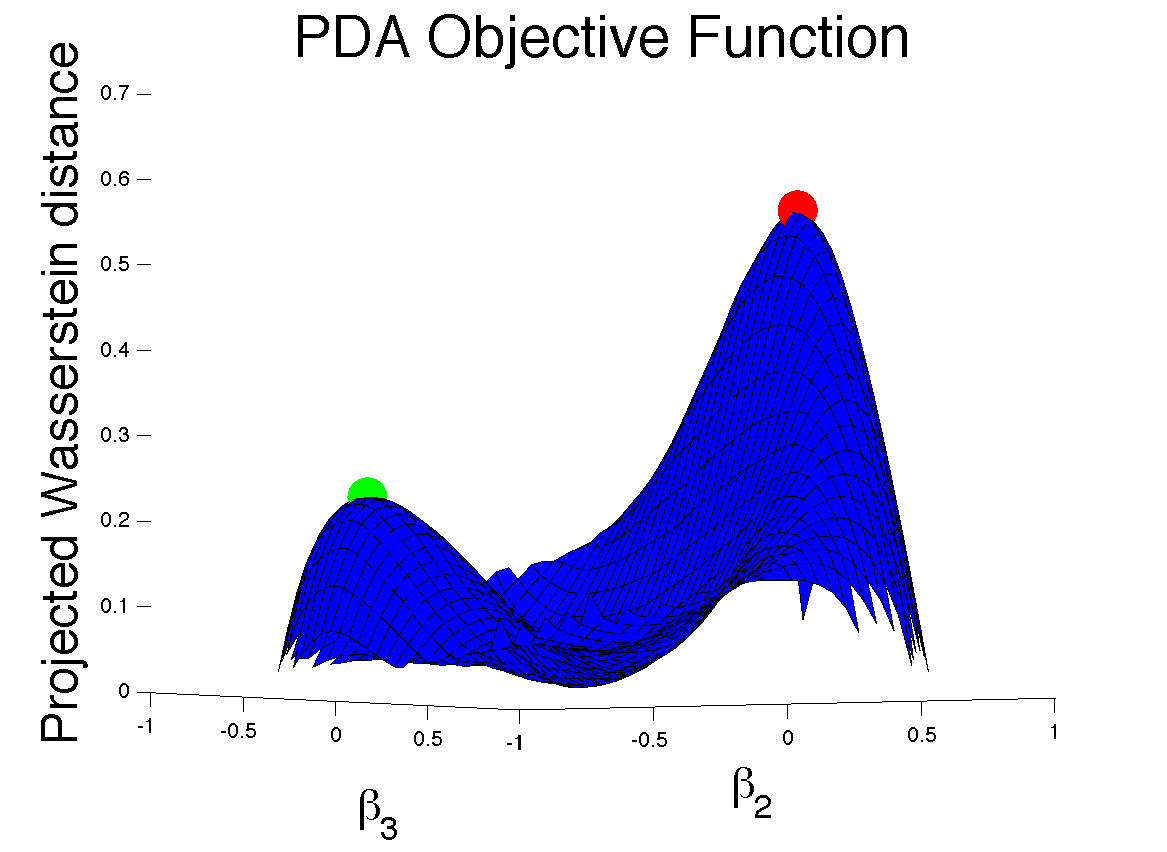} \hspace*{1mm}
\includegraphics[width=0.3\textwidth]{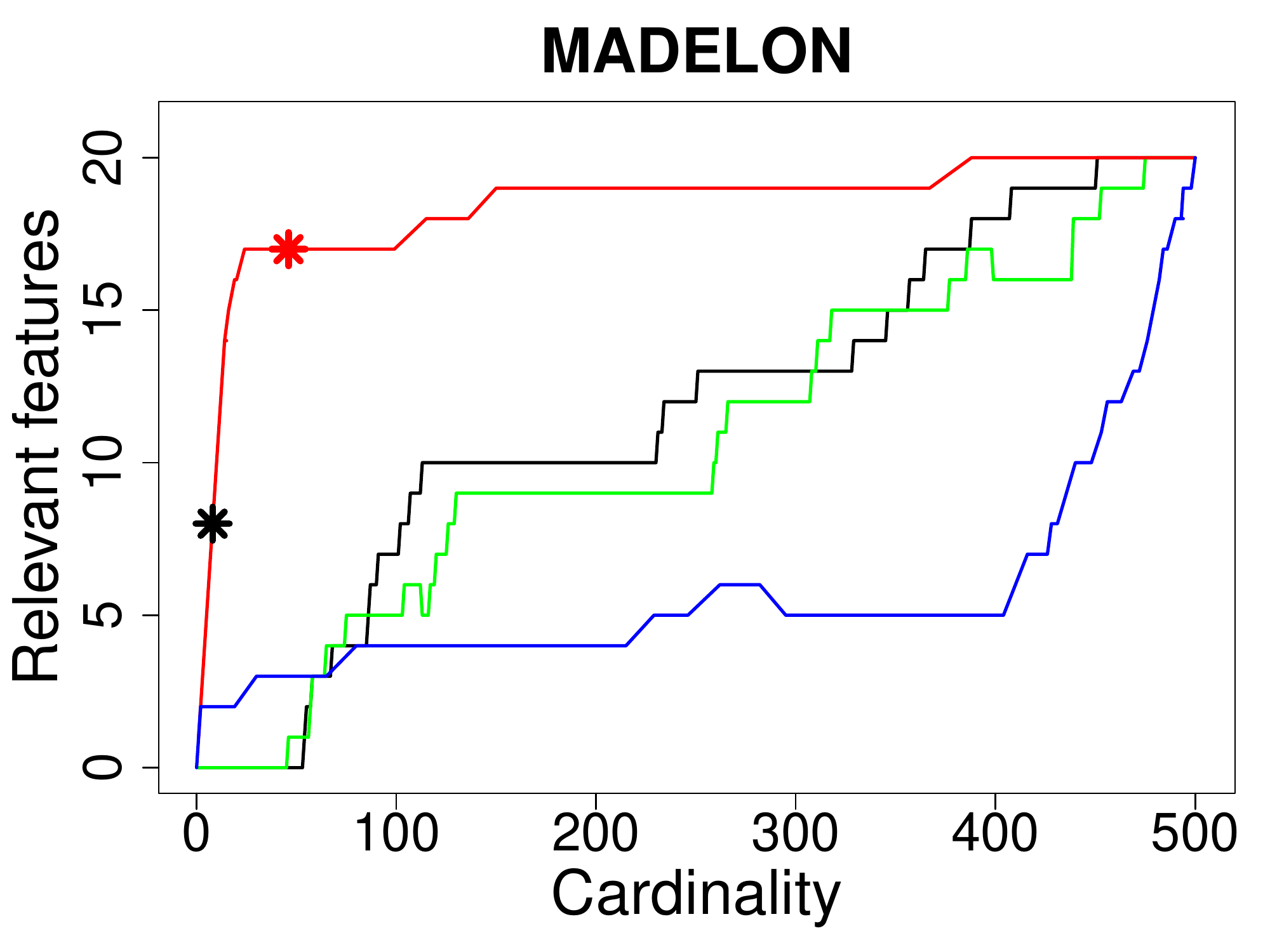}  \hspace*{6mm}
\includegraphics[width=0.3\textwidth]{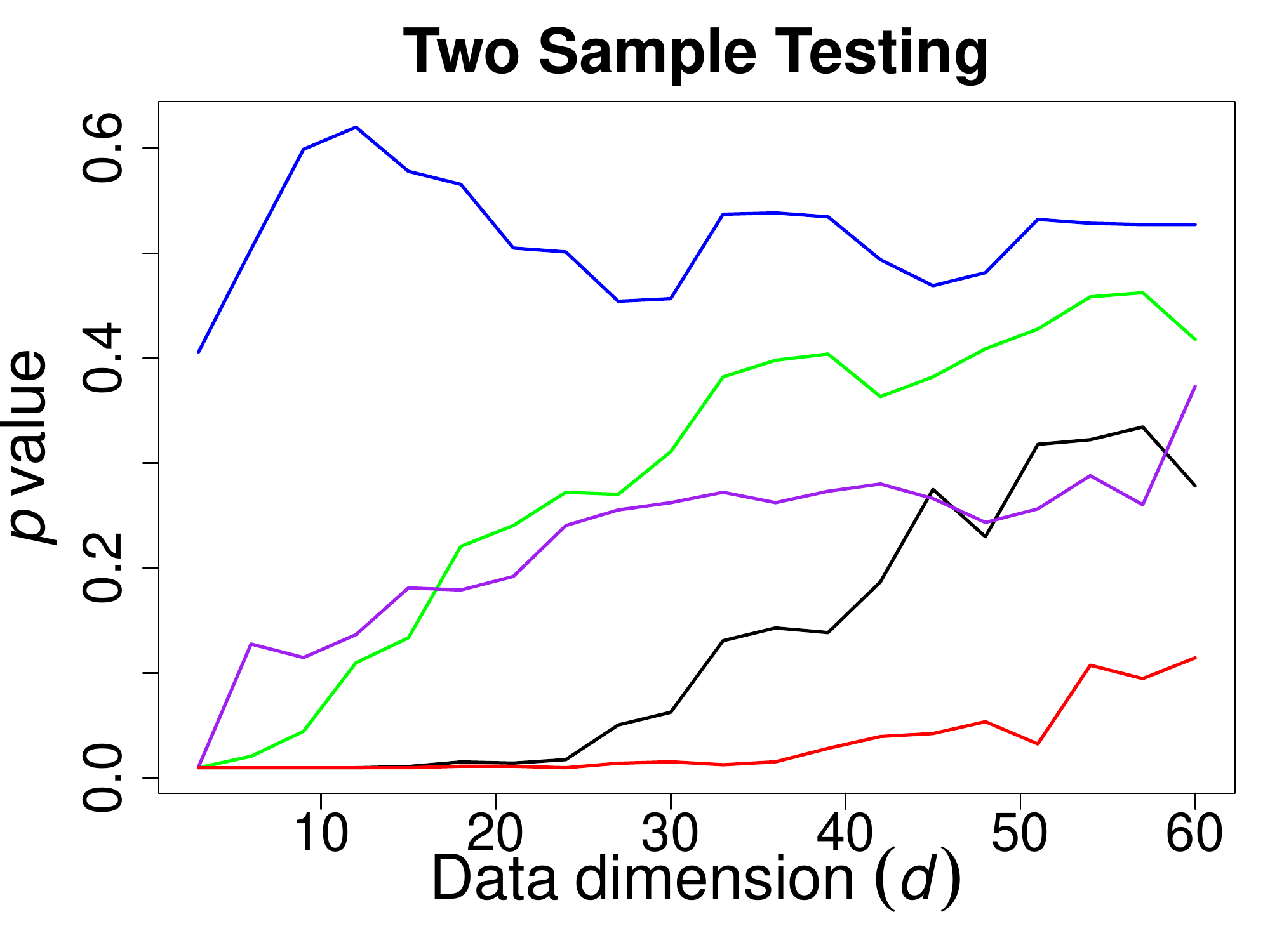} 
\hspace*{13mm} (a) \hspace*{41mm}  (b) \hspace*{43mm} (c) \hspace*{13mm}
\caption[]{(a) example where \methodabbrev{} is nonconvex, (b) \sparsemethod{} vs.\ other feature selection methods, (c) power of various tests for multi-dimensional problems with 3-dimensional differences.} 
\label{spardaresults}
\end{figure}
Figure \ref{spardaresults}a illustrates the cost function of \methodabbrev{} pertaining to two 3-dimensional distributions (see details in Supplementary Information \S \ref{simulate}).  In this example, the point of convergence $\widehat{\beta}$ of the tightening method after random initialization (in green) is significantly inferior to the solution produced by the RELAX algorithm (in red). It is therefore important to use RELAX before tightening as we advise. 

The synthetic MADELON dataset used in the NIPS 2003 feature selection challenge consists of points ($n=m=1000, d= 500$) which have 5 features scattered on the vertices of a five-dimensional hypercube (so that interactions between features must be considered in order to distinguish the two classes), 15 features that are noisy linear combinations of the original five, and 480 useless features  \cite{Guyon2006}.  While the focus of the challenge was on extracting features useful to classifiers, we direct our attention toward more interpretable models. Figure \ref{spardaresults}b  demonstrates how well \sparsemethod{} (red), the top sparse principal component (black) \cite{Zou2005}, sparse LDA (green) \cite{Clemmensen2011}, and the logistic lasso (blue) \cite{tibshirani1996regression} are able to identify the 20 relevant features over different settings of their respective regularization parameters (which determine the cardinality of the vector returned by each method).  The red asterisk indicates the \sparsemethod{}  result with $\lambda$ automatically selected via our cross-validation procedure (without information of the underlying features' importance), and the black asterisk indicates the best reported result in the challenge \cite{Guyon2006}.  

The restrictive assumptions in logistic regression and linear discriminant analysis are not satisfied in this complex dataset resulting in poor performance. Despite being class-agnostic, PCA was successfully utilized by numerous challenge participants \cite{Guyon2006}, and we find that the sparse PCA performs on par with logistic regression and LDA. Although the lasso fairly efficiently picks out 5 relevant features, it struggles to identify the rest due to severe multi-colinearity. Similarly, the challenge-winning Bayesian SVM with Automatic Relevance Determination \cite{Guyon2006} only selects 8 of the 20 relevant features.  In many applications, the goal is to thoroughly characterize the set of differences rather than select a subset of features that maintains predictive accuracy. \sparsemethod{} is better suited for this alternative objective. Many settings of $\lambda$ return 14 of the relevant features with zero false positives. If $\lambda$ is chosen automatically through cross-validation, the projection returned by \sparsemethod{} contains 46 nonzero elements of which 17 correspond to relevant features.

Figure \ref{spardaresults}c depicts (average) $p$-values produced by \sparsemethod{} (red), \methodabbrev{} (purple), the overall  Wasserstein distance in $\mathbb{R}^d$ (black), Maximum Mean Discrepancy \cite{Gretton2012} (green), and DiProPerm \cite{Wei2015} (blue) in two-sample synthetically controlled problems where $\mathbb{P}_X \neq \mathbb{P}_Y$ and the underlying differences have varying degrees of sparsity. Here, $d$ indicates the overall number of features included of which only the first 3 are relevant (see Supplementary Information \S\ref{simulate} for details). As we evaluate the significance of each method's statistic via permutation testing, all the tests are guaranteed to exactly control Type I error \cite{Good1994}, and we thus only compare their respective power in determining $\mathbb{P}_X \neq \mathbb{P}_Y$ setting. The figure demonstrates clear superiority of \sparsemethod{} which leverages the underlying sparsity to maintain high power even with the increasing overall dimensionality. Even when all the features differ (when $d=3$), \sparsemethod{} matches the power of methods that consider the full space despite only selecting a single direction (which cannot be based on  mean-differences as there are none in this controlled data). This experiment also demonstrate that the unregularized \methodabbrev{} retains greater power than DiProPerm, a similar projection-based method \cite{Wei2015}.

Recent technological advances allow complete transcriptome profiling in thousands of individual cells with the goal of fine molecular characterization of cell populations (beyond the crude average-tissue-level expression measure that is currently standard) \cite{Geiler-Samerotte2013}.  We apply \sparsemethod{} to expression measurements of 10,305 genes profiled in 1,691 single cells from the somatosensory cortex and 1,314 hippocampus cells sampled from the brains of juvenile mice \cite{Zeisel2015}.  The resulting $\widehat{\beta}$ identifies many previously characterized subtype-specific genes and is in many respects more informative than the results of standard differential expression methods (see Supplementary Information \S\ref{geneanalysis} for details).  Finally, we also apply \sparsemethod{} to normalized data with mean-zero \& unit-variance marginals in order to explicitly restrict our search to genes whose relationship with other genes' expression is different between hippocampus and cortex cells. This analysis reveals many genes known to be heavily involved in signaling, regulating important processes, and other forms of functional interaction between genes (see Supplementary Information \S\ref{geneinteraction} for details).  These types of important changes cannot be detected by standard differential expression analyses which consider each gene in isolation or require gene-sets to be explicitly identified as features \cite{Geiler-Samerotte2013}.

\section{Conclusion}
This paper introduces the overall principal differences methodology and demonstrates its numerous practical benefits of this approach.  While we focused on algorithms for \methodabbrev{} \& \sparsemethod{} tailored to the Wasserstein distance, different divergences may be better suited for certain applications. 
\nopagebreak

Further theoretical investigation of the \sparsemethod{} framework is of interest, particularly in the high-dimensional $d = O( n)$ setting.  Here, rich theory has been derived for compressed sensing and sparse PCA by leveraging ideas such as restricted isometry or spiked covariance \cite{Amini2009}.  A natural question is then which analogous properties of $\mathbb{P}_X , \mathbb{P}_Y$   theoretically guarantee the strong empirical performance of \sparsemethod{} observed in our high-dimensional applications.  Finally, we also envision extensions of the methods presented here which employ multiple projections in succession, or adapt the approach to non-pairwise comparison of multiple populations.   
\subsubsection*{Acknowledgements}
{\small \vspace*{-2mm} This research was supported by NIH Grant T32HG004947.}
\samepage

\clearpage
\subsubsection*{References}
\begingroup
\renewcommand{\section}[2]{}%
\bibliographystyle{MyOwnBibliographyStyle}
{\small{
\bibliography{DistributionProjectionBibliography}
}}

\endgroup
\clearpage
\newpage \beginsupplement


\begin{center}
{\Huge \textbf{Supplementary Information}}
\end{center}
\begingroup 
\let\orignumberline\numberline
\def\numberline#1{\orignumberline{#1}\kern1ex}
\renewcommand{\baselinestretch}{0.75}\normalsize
\setcounter{tocdepth}{0}
\tableofcontents
\addtocontents{toc}{\setcounter{tocdepth}{2}}
\endgroup
\listoffigures
\listoftables
\renewcommand{\baselinestretch}{1.0}\normalsize

\vspace*{1mm}
\section{Details of simulation study} \label{simulate}

To generate the cost function depicted in  Figure \ref{spardaresults}a, we draw $n = m = 1000$ points from mean-zero 3-dimensional Gaussian distributions with the following respective covariance matrices:
$$ \Sigma_X =       \begin{bmatrix}
    1 & 0.2 & 0.4  \\ 
    0.2 & 1 & 0 \\
    0.4 & 0 & 1 
  \end{bmatrix}  \hspace*{20mm} \Sigma_Y =       \begin{bmatrix}
    1 & -0.9 & 0  \\ 
    -0.9 & 1 & 0 \\
    0 & 0 & 1 
  \end{bmatrix}
$$
Due to the large sample sizes, the empirical distributions accurately represent the underlying populations, and thus the projection produced by the tightening procedure (in green) is significantly inferior to the projection produced by the RELAX algorithm (in red) in terms of actual divergence captured.  Note that only dimensions 2 and 3 of the projection-space are plotted in the figure since {\small $\beta_1 = \sqrt{1 - \sum_{\ell=2}^d \beta_\ell^2}$} is fixed for the unit-norm projections of interest.

Next, we detail the process by which the data are generated for the two-sample problems depicted in Figure \ref{spardaresults}c.  We set the features of the underlying $X$ and $Y$ to  mean-zero multivariate Gaussian distributions in blocks of 3, where within each block, (common) covariance parameters are sampled from the Wishart($\mathbf{I}_{3 \times 3}$) distribution with 3 degrees of freedom.  Only for the first block of 3 features  do we sample a separate covariance matrix for $X$ and a separate covariance matrix for $Y$, so all differences between the two distributions lie in the first 3 features.  To generate a dataset with $d = 3 \times \ell$, we simply concatenate $\ell$ of our blocks together (always including the first block with the underlying difference) and draw $n = m = 100$ points from each class.  We generate 20 datasets by increasing $\ell$ (so the largest $d = 60$), and repeat this entire experiment 10 times reporting the average $p$-values in Figure \ref{spardaresults}c.  

Each $p$-value is obtain by randomly permuting the class labels and recomputing the test statistic 100 times (where we use the same permutations between all datasets).  In SPARDA, regularization parameter $\lambda$ is re-selected using our cross-validation technique in each permutation.  The overall Wasserstein distance in the ambient space is computed by solving a transportation problem \citesi{Levina2001}, and we note the similarity between this statistic and the cross-match test \citesi{Rosenbaum2005}.  A popular kernel method for testing high-dimensional distribution equality, the mean map discrepancy, is computed using the Gaussian kernel with bandwidth parameter chosen by the ``median trick'' \citesi{Gretton2012} (this is very similar to the energy test of \citesi{Szekely2004}).  Finally, we also compute the DiProPerm statistic, employing the the DWD-$t$ variant recommended for testing general equality of distributions \citesi{Wei2015}.

\section{Single cell gene expression in cortex vs.\ hippocampus} \label{geneanalysis}
Playing critical roles in the brain, the somatosensory cortex (linked to the senses) and hippocampal region  (linked to memory regulation and spatial coding) contain a diversity of cell types \citesi{Zeisel2015}.  It is thus of great interest to identify how cell populations in these regions diverge in developing brains, a question we address by applying SPARDA to single cell RNA-seq data from these regions.   Following  \citesi{Trapnell2014}, we represent gene expression by log-transformed FPKM computed from the sequencing read counts\footnote{available in NCBI's Gene Expression Omnibus (under accession GSE60361)}, so values are directly comparable between genes.  Because expression measurements from individual cells are poorer in quality than transcriptome profiles obtained in aggregate across tissue samples (due to a drastically reduced amount of available RNA), it is important to filter out poorly measured genes and we retain a set of 10,305 genes that are measured with sufficient accuracy for informative analysis \citesi{Trapnell2014}.  

\begin{figure}[h!] \centering
\includegraphics[width=\textwidth]{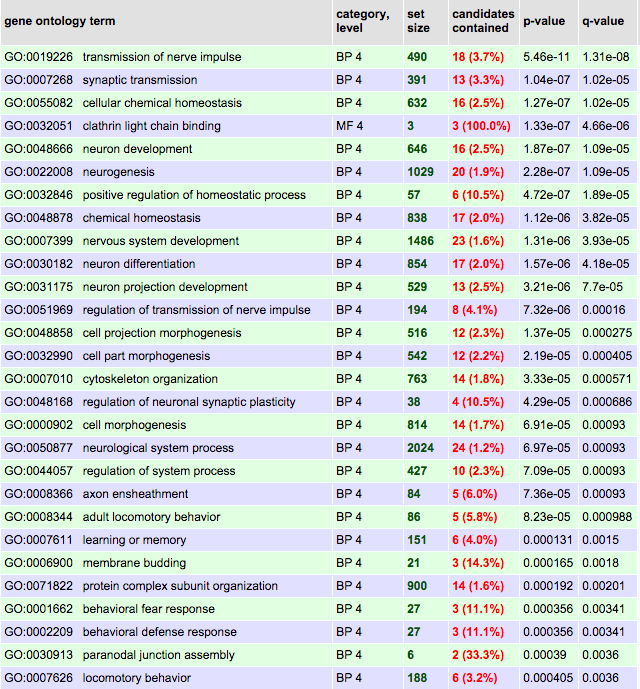}
\caption[Enriched biological processes in annotations of top \sparsemethod{} genes]{Biological process terms most significantly enriched in the annotations of the top 100 genes identified by \sparsemethod{}.}
\label{annotations}
\end{figure}

Table \ref{topgenes} and Figure \ref{annotations} demonstrate that \sparsemethod{} discovers many interesting genes which are already known to play important functional roles in these regions of the brain.  For comparison, we also run LIMMA, a standard method for differential expression analysis which tests for marginal mean-differences on a gene-by-gene basis \citesi{Ritchie2015}.  Ordering the significant genes under LIMMA by magnitude of their mean expression difference, we find that 3 of the top 10 genes identified by \sparsemethod{} also appear in this top 10 list (\emph{Crym}, \emph{Spink8}, \emph{Neurod6}), demonstrating \sparsemethod{}'s implicit attraction toward  large first-order differences over more nuanced changes in practice.  Because only few genes can feasibly be considered for subsequent experimentation in these studies, a good tool for differential expression analysis must rank the most relevant genes very highly in order for researchers to take note.  

\begin{figure}[h!] \centering
\includegraphics[width= 0.6\textwidth]{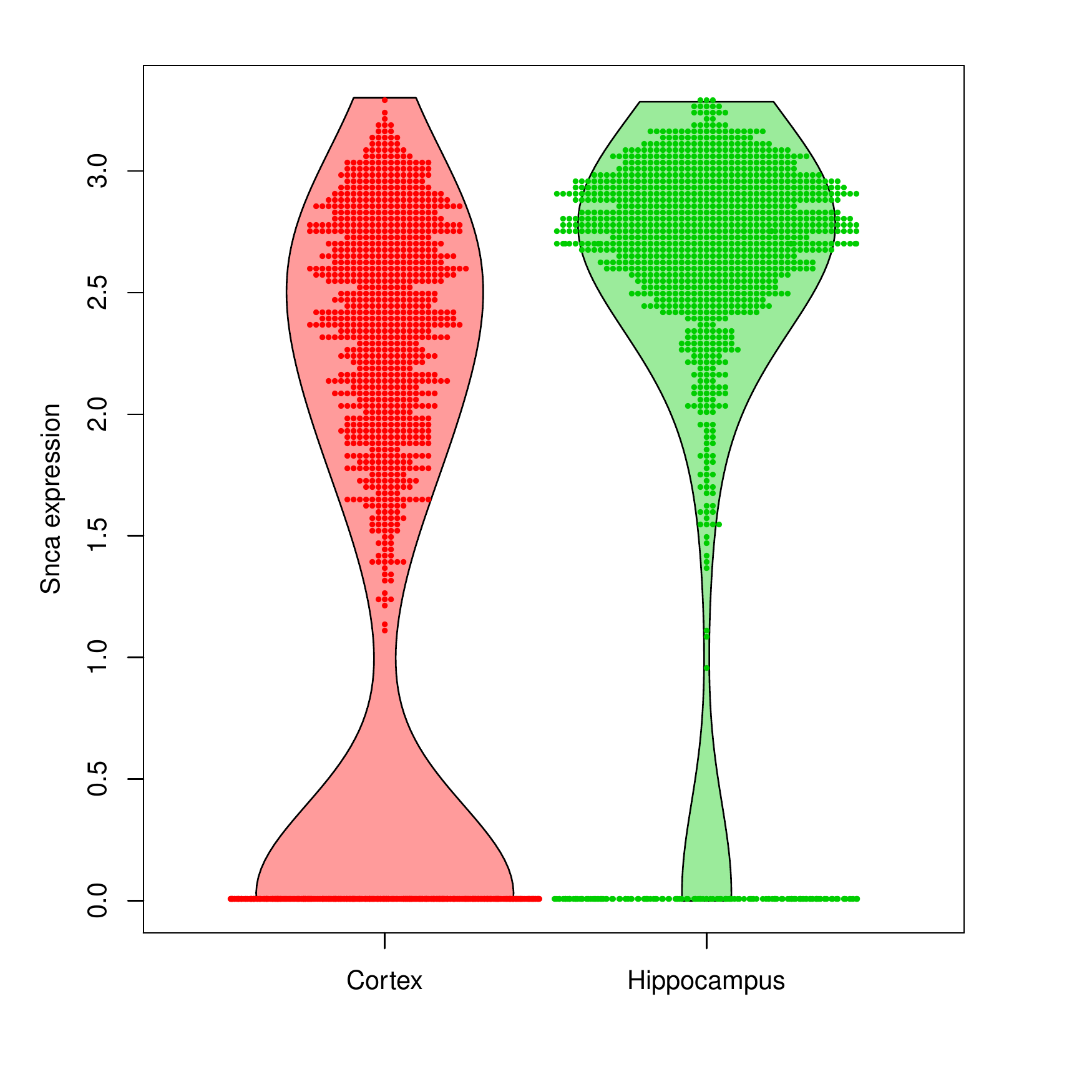}
\caption[Cellular \emph{Snca} expression in somatosensory cortex vs.\ hippocampus]{Distribution of \emph{Snca} expression across cells of the somatosensory cortex and hippocampus.}
\label{snca}
\end{figure}

One particularly relevant gene in this data is \emph{Snca}, a presynaptic signaling and membrane trafficking gene whose defects are implicated in both Parkinson and Alzheimer's disease \citesi{Lesage2009, Linnertz2014}.  While  \emph{Snca} is ranked $11^\text{th}$ highest under \sparsemethod{}, it only ranks 349 according to LIMMA $p$-values and 95 based on absolute mean-shift.  Figure \ref{snca} shows that the primary change in \emph{Snca} expression between the cell types is not a shift in the distributions, but rather the movement of a large fraction of low (1-2.5 log-FPKM) expression cells into the high-expression ($> 2.5$ log-FPKM) regime.  As this type of change does not match the restrictive assumptions of LIMMA's $t$-test, the method fails to highly-rank this gene while the Wasserstein distance employed by \sparsemethod{} is perfectly suited for measuring this sort of effect.  
 
 \begin{table}[h!]
\begin{center}
\begin{tabular}{lcl}
\multicolumn{1}{l}{\bf GENE}  & \multicolumn{1}{l}{\bf WEIGHT}  & \multicolumn{1}{l}{\bf \hspace*{10mm} DESCRIPTION}
\\ \hline \\
Cck         & 0.0593 & Primary distinguishing gene between distinct interneuron classes \\
&& identified in the cortex and hippocampus \citesi{Jasnow2009}   \vspace*{1mm} \\
Neurod6         & 0.0583  & General regulator of nervous system development whose induced mutation \\
& & displays different effects in neocortex vs.\ the hippocampal region \citesi{Bormuth2013}     \vspace*{1mm} \\
Stmn3         &  0.0573  & Up-expressed in hippocampus of patients with depressive disorders \citesi{Oh2010}  \vspace*{1mm}  \\
Plp1         & 0.0570 & An oligodendrocyte- and myelin-related gene which exhibits cortical \\
 & & differential expression in schizophrenia \citesi{Wu2012}  \vspace*{1mm} \\
Crym         & 0.0550  & Plays a role in neuronal specification \citesi{Molyneaux2007} \vspace*{1mm} \\
Spink8         &  0.0536 & Serine protease inhibitor specific to hippocampal pyramidal cells \citesi{Zeisel2015}  \vspace*{1mm} \\
Gap43        & 0.0511 & Encodes plasticity protein important for axonal regeneration \\
& & and neural growth    \vspace*{1mm}  \\
Cryab         &  0.0500 & Stress induction leads to reduced expression in the mouse hippocampus \citesi{Hagemann}   \vspace*{1mm} \\
Mal         &  0.0494 & Regulates dendritic morphology and is expressed at lower levels \\
& &  in cortex than in hippocampus \citesi{Shiota2006}  \vspace*{1mm} \\
Tspan13         &  0.0488 &  Membrane protein which mediates signal transduction events in \\
& & cell development, activation, growth and motility \vspace*{1mm}   \\
\hline
\end{tabular}
\end{center}
\caption[Top genes found by \sparsemethod{}]{Genes with the greatest weight in the projection $\widehat{\beta}$ produced by \sparsemethod{} analysis of the mouse brain  single cell RNA-seq data.  Where not cited, the description of the genes are taken from the standard ontology annotations.}
\label{topgenes}
\end{table}

\subsection{Identifying genes whose interactions differ between  cortex vs.\ hippocampus cells} \label{geneinteraction}
After restricting our analysis to only the top 500 genes with largest average expression (since genes playing important roles in interactions must be highly expressed), we normalize each gene's expression values to have mean zero and unit variance within in the cells of each class.  Subsequent application of SPARDA reveals that most of the genes corresponding to the ten greatest values of the resulting $\widehat{\beta}$ are known to play important roles in in signaling and regulation (see Table \ref{normalized-genes}).

\begin{table}[h!]
\begin{center}
\begin{tabular}{lcl}
\multicolumn{1}{l}{\bf GENE}  & \multicolumn{1}{l}{\bf WEIGHT}  & \multicolumn{1}{l}{\bf \hspace*{10mm} DESCRIPTION}
\\ \hline \\
Thy1         & 0.1245 & Plays a role in cell-cell \& cell-ligand interactions during synaptogenesis \\ 
& & and other processes in the brain \vspace*{1mm} \\
Vsnl1         & 0.1245 & Modulates intracellular signaling pathways of the central nervous system  \vspace*{1mm} \\
Stmn3         & 0.1222  & Stathmins form important protein complex with tubulins  \vspace*{1mm} \\
Stmn2         & 0.1188 & Note: Tubulins Tubb3 and Tubb2 are ranked $20^{\text{th}}$ and $25^{\text{th}}$ by weight in  $\widehat{\beta}$ \ \   \vspace*{1mm} \\
Tmem59         & 0.1176  & Fundamental regulator of neural cell differentiation.  Knock out in the  \\
& & hippocampus results in drastic expression changes of many other genes  \citesi{Zhang2011} \vspace*{1mm} \\
Basp1         & 0.1171 & Transcriptional cofactor which can divert the differentiation of cells to \\
& &     a neuronal-like morphology \citesi{Goodfellow2011}  \vspace*{1mm} \\
Snhg1        & 0.1166 & Unclassified non-coding RNA gene \vspace*{1mm}  \\
Mllt11         & 0.1145  &  Promoter of neurodifferentiation and axonal/dendritic maintenance \citesi{Lederer2007} \vspace*{1mm} \\
Uchl1         & 0.1137 & Loss of function leads to profound degeneration of motor neurons \citesi{Jara2015}. \vspace*{1mm} \\
Cck         & 0.1131  &  Targets pyramidal neurons and enables neocortical plasticity allowing \\
& & for example  the auditory cortex to detect light stimuli \citesi{Li2014, Gallopin2006} \vspace*{1mm}   \\
\hline
\end{tabular}
\end{center}
\caption[Top \sparsemethod{} genes after marginal normalization]{Genes with the greatest weight in the projection $\widehat{\beta}$ produced by \sparsemethod{} analysis of the marginally normalized expression data.}
\label{normalized-genes}
\end{table}

\clearpage
\section{Proofs and Auxiliary Lemmas} \label{sec:proofs}

Throughout this section, we use $C$ to denote absolute constants whose value may change from line to line.  $F$ is defined the cdf of a random variable, and the corresponding quantile function is $F^{-1}(p) := \inf \{ x : F(x) \ge p \}$.  Note our assumptions (A1)-(A3) ensure the quantile function equals the unique inverse of any projected cdf.  Hat notation is used to represent the empirical versions of all quantities, and recall that $D$ denotes the \emph{squared} Wasserstein distance.

\subsection{Proof of Theorem \ref{directionthm}}
\begin{proof}
Since $\widehat{\beta}$ maximizes the empirical divergence, we have: 
\begin{align*}
 & \Pr (  D(\widehat{\beta}^T \widehat{X}^{(n)}, \widehat{\beta}^T \widehat{Y}^{(n)} ) > \Delta - \epsilon) \\
\ge &  \Pr ( D({\beta^*}^T \widehat{X}^{(n)}, {\beta^*}^T \widehat{Y}^{(n)} ) >  \Delta - \epsilon) \\
\ge & \Pr ( D({\beta^*}^T \widehat{X}^{(n)},  {\beta^*}^T X) + D({\beta^*}^T \widehat{Y}^{(n)},  {\beta^*}^T Y) \le \epsilon) \\
\ge & 1 - 4 \exp \left( -\frac{n \epsilon^2}{16 R^4} \right) \text{ applying Lemma \ref{wassbound} and the union bound.}
 \end{align*} 
 \end{proof}

 \begin{lem} For bounded univariate random variable $Z \in [-R , R ]$ with nonzero continuous density in this region, we have 
 \begin{equation*}
 D(\widehat{Z}^{(n)},  Z)  > \epsilon
 \end{equation*}
 with probability at most $ \  2 \exp\left(- \frac{n \epsilon^2}{8 R^4}  \right)  $
 \begin{proof}
On the real line, the (squared) Wasserstein distance is given by:
 \begin{align*}
& D(\widehat{Z}^{(n)},  Z) = \int_0^1 \left( \widehat{F}^{-1}_Z  (p) - F^{-1}_Z (p) \right)^2 \mathrm{d} p \\
& = 4R^2 \int_0^1 \left( \frac{\widehat{F}^{-1}_Z  (p) - F^{-1}_Z (p)}{2R} \right)^2 \mathrm{d} p \ \ \ \text{ where } \ \left| \frac{\widehat{F}^{-1}_Z  (p) - F^{-1}_Z (p)}{2R} \right| \le 1 \ \text{ for each } p \in (0,1) \\
& \le 4 R^2  \int_0^1  \left| \frac{ \widehat{F}^{-1}_Z  (p) - F^{-1}_Z (p)}{2R} \right| \mathrm{d} p \\
& = 2 R  \int_0^1  \left|  \widehat{F}^{-1}_Z  (p) - F^{-1}_Z (p) \right| \mathrm{d} p \\
& =  2R  \int_{-\infty}^\infty \left| \widehat{F}_{Z} (z) - F_{Z} (z) \right|  \mathrm{d} z  \ \  \text{ by the equivalence of 
the (empirical) quantile function and inverse (empirical) cdf } \\
& \le 4R^2 \cdot \sup_z \left| \widehat{F}_{Z} (z) - F_{Z} (z) \right| \\
& \le \epsilon \text{ with probability } \ge 1 - 2 \exp\left(- \frac{n \epsilon^2}{8R^4}  \right)  \text{ by the Dvoretzky-Kiefer-Wolfowitz inequality \citesi{Vaart1996}.} 
  \end{align*} 
 \end{proof}
 \label{wassbound}
 \end{lem}

\clearpage
\subsection{Proof of Theorem \ref{equalitythm}}
\begin{proof} 
We first construct a fine grid of points $\{\alpha_1, \dots, \alpha_S\}$ which form an ($\epsilon/R^2$)-net covering the surface of the unit ball in $\mathbb{R}^d$.  When $\mathbb{P}_X = \mathbb{P}_Y$, the Cramer-Wold device \citesi{Cramer1936} implies that for any point in our grid: $D(\alpha_s^T X, \alpha_s^T Y) = 0$.   A result analogous to Theorem \ref{directionthm} implies $D(\alpha_s^T \widehat{X}^{(n)}, \alpha_s^T \widehat{Y}^{(n)}) > \epsilon$  with  probability $< C_1 \exp\left(-  \frac{C_2}{R^4} n \epsilon^2 \right)$.  

Subsequently, we apply the union bound over the finite set of all grid points.  The total number of points under consideration is the covering number of the unit-sphere which is $\left(1 +  \frac{2R^2}{\epsilon}\right)^d$.
Thus, the probability that $D(\alpha_s^T \widehat{X}^{(n)}, \alpha_s^T \widehat{Y}^{(n)}) < \epsilon$ simultaneously for all points in the grid is at least 
$$ C_1  \left(1 +  \frac{2R^2}{\epsilon}\right)^d \exp\left(-  \frac{C_2}{R^4} n \epsilon^2 \right)
$$

 By construction, there must exist grid point $\alpha_0$ such that  $||\widehat{\beta} - \alpha_0||_2 < \epsilon/R^2$.  By Lemma \ref{lem:close}
\begin{align*}
& D(\widehat{\beta}^T \widehat{X}^{(n)} , \widehat{\beta}^T \widehat{Y}^{(n)} ) \le D( {\alpha_0}^T \widehat{X}^{(n)} ,  {\alpha_0}^T \widehat{Y}^{(n)} ) + C\epsilon 
\end{align*}
thus completing the proof.
\end{proof}
\vspace*{6mm}
\begin{lem} For $\alpha, \beta \in \mathcal{B}$ such that $||\alpha - \beta||_2 < \epsilon$, we have:
\begin{equation}
| D(\alpha^T \widehat{X}^{(n)} , \alpha^T \widehat{Y}^{(n)} ) - D( \beta^T \widehat{X}^{(n)} ,  \beta^T \widehat{Y}^{(n)} ) | \le C  \epsilon R^2
\label{empdistclose}
\end{equation}
\begin{proof} We assume that the $\alpha$-projected divergence is larger than the $\beta$-projected divergence, and write:
$$ D(\beta^T \widehat{X}^{(n)} , \beta^T \widehat{Y}^{(n)} ) =  \min_{M \in \mathcal{M}, } \ \sum_{i=1}^n \sum_{j = 1}^m (\beta^T x^{(i)} -  \beta^T y^{(j)})^2 M_{ij} 
$$
recalling that $\mathcal{M}$ is the set of  matching matrices defined in the main text.  Let $M(\beta)$ denote the matrix which is used in the computation of the $\beta$-projected empirical Wasserstein distance (the minimizer of the righthand side of the above expression). 
Thus, we can express (\ref{empdistclose}) as:
\begin{align*}
& \sum_{i=1}^n \sum_{j = 1}^m (\alpha^T x^{(i)} -  \alpha^T y^{(j)})^2 M(\alpha)_{ij}  - \ \sum_{i=1}^n \sum_{j = 1}^m (\beta^T x^{(i)} -  \beta^T y^{(j)})^2 M(\beta)_{ij} \\
\le & \sum_{i=1}^n \sum_{j = 1}^m (\alpha^T x^{(i)} -  \alpha^T y^{(j)})^2 M(\beta)_{ij}  - \ \sum_{i=1}^n \sum_{j = 1}^m (\beta^T x^{(i)} -  \beta^T y^{(j)})^2 M(\beta)_{ij} \\
\le & \sum_{i=1}^n \sum_{j = 1}^m \left[ (\alpha^T (x^{(i)} - y^{(j)}))^2  - (\beta^T( x^{(i)} -  y^{(j)}))^2 \right] M(\beta)_{ij} \\
= & \sum_{i=1}^n \sum_{j = 1}^m \left[ (\alpha-\beta)^T (x^{(i)} - y^{(j)}) \cdot (\alpha+\beta)^T( x^{(i)} -  y^{(j)}) \right] M(\beta)_{ij} \\
\le & \sum_{i=1}^n \sum_{j = 1}^m C \epsilon R^2 M(\beta)_{ij} = C \epsilon R^2 
\end{align*} 
\end{proof}
\label{lem:close}
\end{lem}
 \clearpage
 
 \subsection{Proof of Theorem \ref{multivariatethm}}
 \begin{proof} Our proof relies primarily on a quantitative form of the Cramer-Wold result presented in \citesi{Jirak2011}.  We adapt Theorem 3.1 \citesi{Jirak2011} in its contrapositive form:  If $\exists \ a \ge 0$ such that $T_a( X ,  Y ) > h( g(\Delta))$, then $\exists \beta \in \mathcal{B}$ such that 
 \begin{equation} \sup_{z \in \mathbb{R}} \ \ \bigg|  \Pr\left(  \beta^T X \le z  \right)  -  \Pr\left(  \beta^T Y \le z  \right)  \bigg| > g( C \Delta)
 \label{kolm}
\end{equation}
 Subsequently we leverage a number of well-characterized relationships between different probability metrics (cf.\ \citesi{Gibbs2002}) to lower bound the projected (squared) Wasserstein distance (of the underlying random variables).
 
Letting $K_\beta$ denote the projected Kolmogorov distance in (\ref{kolm}), we have that the $\beta$-projected L\'{e}vy-distance, $L_\beta$ satisfies:
\begin{equation}
K_\beta \le (1 + \Phi) L_\beta \ \ \ \ \text{ where } \Phi := \sup_{\alpha \in \mathcal{B}} \ \big\{ \sup_y |f_{\alpha^T Y}(y) | \big\}
\end{equation}
and $f_{\alpha^T Y}(y)$ is the density of the projection of $Y$ in the $\alpha$ direction.

In turn the projected L\'{e}vy $L_\beta$ is bounded above by the Prokhorov metric which itself is bounded above by the square root of the $\beta$-projected Wasserstein distance.   Following the chain of inequalities, we obtain: $D(\beta^T X , \beta^T Y) \ge  C \Delta$, to which we can apply Theorem \ref{directionthm} to obtain the desired probabilistic bound on the empirical projected divergence.
\end{proof}
\vspace*{5mm}
\subsection{Proof of Theorem \ref{sparsistency}}
  \begin{proof} Theorem \ref{equalitythm} implies that with high probability, any $\beta_{S^C} \in \mathbb{R}^{d-k}$ has
  $D(\beta_{S^C}^T  \widehat{X}^{(n)}_{S^C} , \beta_{S^C}^T \widehat{Y}^{(n)}_{S^C} ) < \epsilon$, while Theorem \ref{multivariatethm} specifies the probability that there exists $\beta_{S} \in  \mathbb{R}^{k}$ such that $D(\widehat{\beta}_S^T \widehat{X}^{(n)}_S, \widehat{\beta}_S^T \widehat{Y}^{(n)}_S ) > d \cdot \epsilon$.
  
A bound for the probability that both theorems' conclusions hold may be obtained by the union bound.  When this is the case, it is clear that the optimal $k$-sparse $\widehat{\beta} \in \mathbb{R}^d$ must obey the sparsity pattern specified in the statement of Theorem \ref{sparsistency}.  To see this, consider any $\beta \in \mathcal{B}$ with $\beta_j \neq 0$ for some $j \in S^C$ and note that it is always possible to produce a strictly superior projection by setting $\beta_j = 0$ and distributing the additional  weight $| \beta_j |$ among the features in $S$ in an optimal manner.
 \end{proof}
 \vspace*{10mm}
 
\section{Derivation of semidefinite relaxation properties} \label{relaxderivations}
Here, we provide some intuitive arguments for the conclusions in \S\ref{relaxation-property}, regarding some conditions under which our semidefinite relaxation is nearly tight.

Condition (i) derives from the fact that (\ref{relaxation}) has rank one solution when the objective is approximately linear in $B$.  

(ii) and (iii) are derived by noting that (\ref{relaxation}) is the Wasserstein distance between random variables $B^{1/2} X$ and $B^{1/2} Y$ where $A X$ follows a $N(A \mu_X, A \Sigma_X A^T)$ distribution when $X$ is Gaussian. Furthermore, the Wasserstein distance between (multivariate) Gaussian distributions can be analytically written as
$$W(X,Y) =  ||\mu_X - \mu_Y||_2^2  +  || \Sigma_X^{1/2} - \Sigma_Y^{1/2}  ||_F^2 $$
 
\newpage   \begin{changemargin}{-0.2cm}{-0.2cm}
\bibliographystylesi{MyOwnBibliographyStyle} 
{\small
\bibliographysi{DistributionProjectionBibliography}
}
\end{changemargin}

\end{document}